\documentclass[10pt,twocolumn,letterpaper]{article}

\usepackage{wacv}
\usepackage{times}
\usepackage{epsfig}
\usepackage{graphicx}
\usepackage{amsmath}
\usepackage{amssymb}

\usepackage{multirow}
\usepackage{array}
\usepackage[export]{adjustbox}

\def\wacvPaperID{3} 

\wacvfinalcopy 

\ifwacvfinal
\def\assignedStartPage{9876} 
\fi
\ifwacvfinal
\usepackage[breaklinks=true,bookmarks=false]{hyperref}
\else
\usepackage[pagebackref=true,breaklinks=true,colorlinks,bookmarks=false]{hyperref}
\fi

\ifwacvfinal
\else
\fi

\begin{document}

\title{A Log-likelihood Regularized KL Divergence for Video Prediction With a 3D Convolutional Variational Recurrent Network}

\author{Haziq Razali and Basura Fernando \\
A*STAR, Singapore\\
{\tt\small \{Haziq\_Razali, Basura\_Fernando\}@ihpc.a-star.edu.sg}
}


\maketitle

\begin{abstract}
The use of latent variable models has shown to be a powerful tool for modeling probability distributions over sequences. In this paper, we introduce a new variational model that extends the recurrent network in two ways for the task of video frame prediction. First, we introduce 3D convolutions inside all modules including the recurrent model for future frame prediction, inputting and outputting a sequence of video frames at each timestep. This enables us to better exploit spatiotemporal information inside the variational recurrent model, allowing us to generate high-quality predictions. Second, we enhance the latent loss of the variational model by introducing a maximum likelihood estimate in addition to the KL divergence that is commonly used in variational models. This simple extension acts as a stronger regularizer in the variational autoencoder loss function and lets us obtain better results and generalizability. Experiments show that our model outperforms existing video prediction methods on several benchmarks while requiring fewer parameters.
\end{abstract}

\section{Introduction}
Generating the future frames given the past has been a long standing problem in Computer Vision. Currently, recurrent neural networks, specifically a variant with Long Short Term Memory (LSTM) cells \cite{lstm}, hold the state-of-the-art results in a wide range of sequence based tasks including future frame prediction \cite{villegas2019high,lee2018stochastic,wang-predrnn,wang-e3d}. At a high level, they belong to the family of autoregressive models where the predicted element is conditioned on the history of inputs received thus far. From video analysis \cite{videoanalysis1} to speech recognition \cite{speechrecog1}, text generation \cite{textgeneration1}, machine translation \cite{machinetranslation1} and image captioning \cite{imagecaptioning1}, the versatility of recurrent networks has proven to be an indispensable tool for machine learning practitioners.

There  is  evidence  that  the  introduction  of  uncertainty  into  the  hidden  states of  a  recurrent  network  can  significantly  improve  its  performance  when  modelling complex sequences such as speech and music \cite{storn,vrnn,srnn,zforcing}. These methods integrate the Variational Autoencoder (VAE) \cite{vae} to infer the latent variables which is shown to capture some form of semantic abstraction such as the thickness or orientation of an MNIST digit from the observed data by better capturing the input distribution.

Despite so, the state-of-the-art in future frame prediction using recurrent networks have come from purely deterministic auto-encoder type models \cite{wang-predrnn,wang-e3d}. We believe this to be the consequence of 2 factors: (a) that the use of latent random variables to model sequential data for generation often result in blurry reconstructions as evidenced in the literature \cite{blurry}, perhaps due to the lack of a proper temporal model and (b) that the regularizer employed is not sufficient for capturing the properties of the encoded distribution.
In particular, a VAE is an autoencoder where the training process is regularized to ensure that the latent space captures the input distribution accurately. This regularization is usually employed by minimizing the Kullback-Liebler (KL) divergence between the encoded posterior and prior distributions.
Here, the posterior is assumed to be a standard Gaussian distribution and the encoder is then trained to return the mean and the covariance of the posterior Gaussian distribution. 
to the future).
In this paper, we address the above mentioned problems and propose a new architecture for the prediction of future frames. We extend the Variational Recurrent Neural Network (VRNN) \cite{vrnn} firstly by replacing all fully-connected and 2D convolutions in the architecture with 3D convolutions to increase its capacity to model temporal information. We use a truncated ResNet \cite{resnet,3dresnet} with 3D convolutions for both the image encoder and decoder, a shallow 3D convolutional network to generate the prior and posterior distributions and, a 3D Convolutional LSTM (ConvLSTM) \cite{convlstm} as the recurrent model. Since the architecture is fully fitted with 3D convolutions that share parameters across both space and time, it can now better exploit spatiotemporal dependencies and more importantly, preserve temporal information across each component and operation thereby removing the LSTM’s complete dependency on the hidden states for motion information. Similarly, using 3D convolutions in the image decoder allows it to generate high quality predictions by considering the spatiotemporal correlations of the 4D feature maps in contrast to 2D convolutions. 
Finally, we choose to truncate our 3D-ResNet in order to leverage its power as a spatiotemporal feature extractor while minimizing the total number of parameters. 
We argue that features extracted by the lower layers of a 3D-ResNet are already abstract enough for the ConvLSTM to further learn on and would therefore rather reallocate the freed space to the ConvLSTM. This is in contrast to \cite{hierarchyvrnn} that uses the full 2D-ResNet for feature extraction. On the other hand, we want to avoid forgoing the image encoder as in \cite{wang-predrnn,wang-e3d} since it puts too much reliance on the ConvLSTM to jointly learn short-term spatiotemporal features and long term dynamics from a raw sequence of images.

Next, we further regularize the latent space of the variational recurrent model using a novel latent loss that combines the KL divergence and the log-likelihood criterion. Specifically, we further constraint the latent space by maximizing the likelihood of the prior mean with respect to the posterior distribution assuming that conditionally, the prior given the posterior also follows a normal distribution. We will mathematically show in section \ref{sec:loglikelihood} that this additional constraint lets the prior variance be larger while reducing the divergence between the prior and posterior mean distributions. Ultimately, we will show in our experiments that our novel objective function combined with our architectural design choice lets us outperform the state of the art while requiring fewer parameters.

In summary, we make three contributions. First, we present a VRNN that uses 3D convolutions across the entire architecture and show their effectiveness for future frame prediction. Second, we extend the KL divergence by introducing a novel log-likelihood criterion to the latent loss used in variational models. This new loss further regularizes the latent space and allows us to obtain better results.
Finally, we show through experiments that each individual contribution improves the model and that their combination allows the model to outperform existing state-of-the-art video prediction methods.

\section{Related Work}
\label{sec:relatedwork}
Recurrent Networks used to predict the future frames can be grouped into two categories: (1) those that are entirely deterministic and (2) those that propagate uncertainty through the recurrent network via latent random variables.

Recurrent networks were first used for future frame prediction in \cite{ranzato} when Ranzato et al. learnt a model to predict a quantized space of image patches. 
Srivastava et al. \cite{srivastava} proposed a model to predict the future as well as the input sequence in order to prevent the model from storing information only about the last few frames.
Shi et al. \cite{convlstm} proposed an extension of the LSTM by replacing the fully connected structure with one that is fully convolutional which saw popular use to date for learning sequential data with spatial information. 
Finn et al. \cite{finn} used an LSTM framework to model motion via transformations of groups of pixels. 
Patraucean et al. \cite{patraucean} and Villegas et al. \cite{villegas} explicitly injected short term motion information through the use of optical flow. 
Xu et al. \cite{xu} proposed a two-stream recurrent network to deal with the high and low frequency content often present in natural videos. 
Kalchbbrenner et al. \cite{kalchbrenner} introduced a model that learns the joint distribution of the raw pixels to generate them one at a time. 
Wang et al. \cite{wang-predrnn} proposed to improve the stacked LSTM by having the memory and hidden states flow in a zig-zagged manner from the highest unit of the current timestep to the lowest unit of the subsequent timestep. 
This was further improved in \cite{wang-e3d} by replacing the 2D convolutions with 3D and a memory attention in the LSTM itself. 
We also use 3D convolutions throughout our entire architecture but in contrast, our model is stochastic.

Stochastic recurrent networks vary in way they propagate uncertainty across time as well as the way inference is computed. For instance, Bayer et al. \cite{storn} and Goyal et al. \cite{zforcing} conditioned the generation only on the hidden states of the recurrent network whilst Chung et al. \cite{vrnn} and Fraccaro et al. \cite{srnn} have the output be some function over both the hidden states and the latent vector. Next, the LSTM state transitions in \cite{storn,vrnn,zforcing} are additionally conditioned on the latent vectors whereas in \cite{srnn} is not. The work of \cite{vrnn} was later extended in \cite{hierarchyvrnn} through a hierarchy of latent variables for future frame prediction. We propagate stochastic information in the same way as \cite{vrnn} except that the latent tensors themselves now contain richer spatial-temporal information since they are the result of 3D convolutions.
For inference, both \cite{srnn} and \cite{zforcing} run a deterministic recurrent network backwards through the sequence to form the approximate posterior whereas the posterior in \cite{storn} and \cite{vrnn} is computed using only information up till the present. 
Similarly, our method for inference follows that of \cite{vrnn} but in contrast to \cite{vrnn} and in fact all existing methods, we jointly optimize both the KL divergence and a novel log likelihood criterion.

Future frame prediction is useful for various application such as early action prediction\cite{Ryoo2011,shi2018action,sadegh2017} and action anticipation~\cite{Furnari2019,Miech2019,qi2017predicting}.
Some anticipation methods generate future RGB images and then classify them into human actions using convolution neural networks ~\cite{Zeng2017a,Wang2017}. 
However, generation of RGB images for the future is a very challenging task specially for longer action sequences.
Similarly, some methods aim to generate future motion images and then try to predict  action for the future~\cite{Rodriguez2018}.

\section{Our model}
Given a sequence of frames $\textbf{x}_{1:C-1}$ as context, our goal is to learn a model that can predict $T$ frames into the future i.e. $\hat{\textbf{x}}_{C:C+T}$. 
This task is challenging due to the variability present in video sequences and the fact that there can exist multiple plausible futures for any given input. To overcome this, we propose to use a VRNN but with 3D convolutions in order to better capture both short and long term relations.
Specifically, in contrast to existing VRNN models, our encoder, decoder, prior and posterior networks, and LSTM are all built using 3D convolutions and up-convolutions. We also further regularize the latent space of the VRNN with an additional log-likelihood term. We begin the next section with a brief review of the VRNN before describing our novel latent loss function and the architecture of our 3D VRNN.

\subsection{Variational Recurrent Neural Network}
\label{sec:variationalconvlstm}
\begin{figure}
\begin{center}
\begin{tabular}{c}
\includegraphics[width=0.6\columnwidth]{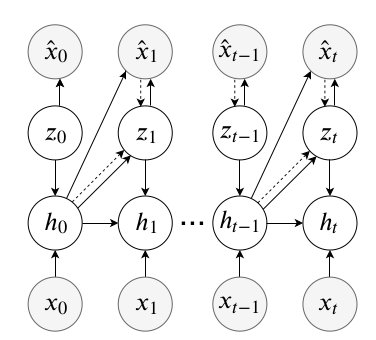}
\end{tabular}
\caption{Graphical illustration of the Variational Recurrent Network. The dotted lines denote the posterior network $\text{f}_{\text{q}}$ that is only used during training and is discarded at test time.}
\label{fig:vrnn}
\end{center}
\end{figure}

Figure \ref{fig:vrnn} provides a graphical illustration of the VRNN. 
The VRNN uses a latent variable $\textbf{z}_{t}$ at each timestep of a recurrent network to capture the variations in the observed data. 
It contains a VAE at every timestep whose mean $\mu_{t}$ and variance $\sigma_{t}$ are conditioned on the hidden unit $h_t$ of a recurrent network. These parameters are then used to sample the latent variable $\textbf{z}_{t}$ at each timestep. 
Concisely, the forward pass can be completely described by the following set of recurrence equations where the subscripts p and q denote the prior and posterior distributions respectively, and the components $\text{f}_{\text{p}}$, $\text{f}_{\text{q}}$, $\text{f}_{\text{enc}}$, $\text{f}_{\text{dec}}$ are functions implemented using neural networks.

\begin{align}
    \mu_{\text{p},t},\sigma_{\text{p},t} &= \text{f}_{\text{p}}(h_{t-1})\\ 
    \mu_{\text{q},t},\sigma_{\text{q},t} &= \text{f}_{\text{q}}(h_{t-1}, \text{f}_{\text{enc}}(x_t))\\
    z_{\text{p},t} &\sim N(\mu_{\text{p},t},\sigma_{\text{p},t})\\
    z_{\text{q},t} &\sim N(\mu_{\text{q},t},\sigma_{\text{q},t})\\
    \hat{x}_t &= \text{f}_{\text{dec}}(z_{\text{p},t}, h_{t-1})\\
    h_{t} &= \text{LSTM}(\text{f}_{\text{enc}}(x_t),h_{t-1},z_{\text{p},t})
\end{align}
Here $N(\mu,\sigma)$ is a multivariate Gaussian distribution with mean $\mu$ and co-variance \textbf{diag}$(\sigma^2)$. 
Note that the posterior network $\text{f}_{\text{q}}$ is used only during training and is discarded at test time. 
The entire model is then trained end-to-end for future frame prediction by minimizing a sum of the reconstruction loss ($L_{\text{rec}}$), and latent loss ($L_{\text{latent}}$) expressed as:
\begin{align}
	L = \lambda_{\text{rec}} L_{\text{rec}} + \lambda_{\text{latent}} L_{\text{latent}}
\end{align}
where $\lambda_{\text{rec}}$ and $\lambda_{\text{latent}}$ are the trade off hyper-parameters and the latent loss is the timestep-wise KL divergence ($L_{\text{KL}}$) between the prior ($p$) and posterior ($q$) distributions and is expressed as: 
\begin{align}
     & L_{\text{KL}} = \sum_{t=1}^T \text{KL}(q(z_t | X_{\le t}, Z_{< t}) || p(z_t | X_{< t}, Z_{< t})))= \\
    & \sum_{t=1}^T \log (\sigma_{\text{q,t}}) - \log (\sigma_{\text{p,t}}) +  \frac{\sigma_{\text{p,t}}^2 +  (\mu_{\text{p,t}}-\mu_{\text{q,t}})^2}{2\sigma_{\text{q,t}}^2} - 0.5 
\label{eq:kldenominator}
\end{align}


\begin{figure*}[t!]
\begin{center}
\begin{tabular}{c}
\includegraphics[width=1\textwidth]{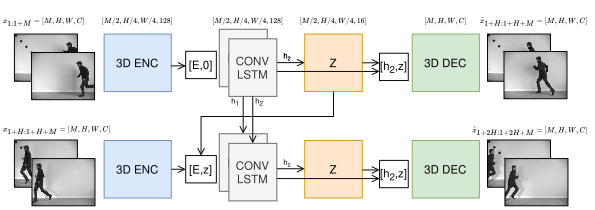}
\end{tabular}
\caption{Our proposed architecture for future frame prediction. The architecture inputs and outputs at each timestep a sequence of M video frames with the prediction made H timesteps into the future. The entire architecture is fitted with 3D convolutions. The 3D-ENC and 3D-DEC are mirrored versions of the 2-block 3D-ResNet18 as shown in Figure \ref{fig:encoderdecoder}. We use 2 LSTM layers and a shallow 3D conv network to generate $\mu$ and $\sigma$ that are then sampled from to produce z. The block $[\circ,\circ]$ indicates a concatenation along the 4th axis. The values above each component (3D-ENC, CONVLSTM, Z, 3D-DEC) indicate the sizes of the output tensor.}
\label{fig:architecture}
\end{center}
\end{figure*}

\subsection{New log-likelihood regularized KL divergence}
\label{sec:loglikelihood}
Typically, the KL divergence-based latent loss is used to regularize the latent space, enforcing it to be a Gaussian distribution with known parameters. 
We further enhance this regularization by appending the negative of the log-likelihood term to the latent loss. 
The objective here is to maximize the likelihood of the prior mean distribution w.r.t. the posterior. 
This is done by minimizing the negative likelihood as shown in Eq.~\ref{eq:lldenominator} by assuming that the prior, posterior and \emph{the conditional prior mean} given the posterior all follow a Gaussian distribution.
\begin{align}
   - L_{\text{LL}} &= - \log \prod_{t=1}^T p(\mu_{\text{p},t} | \mu_{\text{q},t}, \sigma_{\text{q},t}) \\
    &= \sum_{t=1}^T \log ({\sigma_{\text{q},t}}) + ( \frac{\mu_{\text{p},t} - \mu_{\text{q},t}}{\sigma_{\text{q},t}} )^2
    \label{eq:lldenominator}
\end{align}

The proposed latent loss is thus expressed together as:

\begin{align}
& L_{\text{KL}} - L_{\text{LL}} \nonumber \\
&= \log (\sigma_{\text{q}}) - \log (\sigma_{\text{p}}) + \frac{\sigma_{\text{p}}^2 + (\mu_{\text{p}}-\mu_{\text{q}})^2}{2\sigma_{\text{q}}^2} - 0.5 \nonumber \\
&+ \log ({\sigma_{\text{q}}}) + ( \frac{\mu_{\text{p}} - \mu_{\text{q}}}{\sigma_{\text{q}}} )^2 \\
&= 2\log (\sigma_{\text{q}}) - \log (\sigma_{\text{p}}) + \frac{\sigma_{\text{p}}^2 + 3(\mu_{\text{p}}-\mu_{\text{q}})^2}{2\sigma_{\text{q}}^2} - 0.5
\end{align}

Interestingly, the above equation is similar to the KL divergence (eq \ref{eq:kldenominator}) but with some of its components weighted differently. In particular, the log posterior variance, $\log (\sigma_q)$,  has been scaled by a factor of 2 and the squared difference of mean, $(\mu_p - \mu_q)^2$, by a factor of 3. This modification has 2 effects. First, it puts more emphasis on sample diversity since the log prior variance $\log(\sigma_p)$ put out by the network must now be higher in order to match the scaled log posterior variance $2\log(\sigma_q)$. Secondly, the scaled difference of mean $3(\mu_p - \mu_q)^2$ serves to balance out the additional weight assigned to the variance term and thus encourages the model to continue generating samples that are representative of the dataset. As such, the log-likelihood regularized KL divergence should have no adverse effects on the model since it is simply the KL divergence with a reweighting of its components and would argue it to be more forceful if one needs to have a greater emphasis on sample diversity. 
Interestingly, the weights for each component can also be customized although their individual effects will not be investigated since it is not the purpose of this paper.
All-in-all, our new loss function for training the VRNN is expressed together as: $L = \lambda_{\text{rec}} L_{\text{rec}} + \lambda_{\text{latent}} ( L_{\text{KL}} - L_{\text{LL}} )$. 

\subsection{Our 3D Convolutional VRNN}
\label{sec:convolutionalrecurrentnetworks}
The ConvLSTM was proposed in \cite{convlstm} to address the shortcomings of Fully-Connected LSTM, namely that latter always ends up decimating any spatial information contained in the input tensor. 
%
Intuitively, if the states are viewed as the hidden representations of moving objects, then a ConvLSTM with a larger kernel should be able to capture faster motions while one with a smaller kernel can capture slower motions.
However, if the input at each timestep is a single image, then the hidden states are the only component that carry motion information in both Fully-Connected and ConvLSTMs.


In our work, we counteract this limitation by replacing all 2D convolutions (and de-convolutions) with 3D to enable every component to retain motion information instead of only the hidden states. The benefits of this are two-fold. First, the 3D ConvLSTM is no longer completely reliant on the hidden states for motion information since there is an additional source coming from the 3D image encoder. Specifically, the use of 3D convolutions on multiple frames result in an input tensor that carries short-term spatiotemporal information as opposed to a VRNN that run a 2D convolution on a single frame at every timestep. Second, we can now vary the window size and output horizon at each timestep without needing to redesign the architecture. For example, let us define \textbf{M} to be the window size, or the number of input frames to our model at each timestep and $\mathcal{H}$ the output horizon (output frames), or how far into the future should the model predict. Then, 3D convolutions (and de-convolutions) allow us to set a large \textbf{M} to efficiently capture large motions when dealing with datasets where the motion between frames is prevalently large and conversely, a large $\mathcal{H}$ to predict many frames into the future at once with minimal reconstruction errors if said motion between frames is small. All in all, this upgrade renders our 3D convolutional VRNN more effective and general than its 2D counterpart. 

Our proposed architecture is shown in Figure \ref{fig:architecture}. The encoder (3D-ENC) takes at each timestep a clip of M video frames of shape [M,H,W,C] to produce a tensor of shape [M/2, H/4, W/4, C/4] where H,W,C denote the height, width and channels respectively. This tensor is then concatenated with the latent tensor Z (a zero tensor in the 1st timestep) along the 4'th channel indicated by $[\circ,\circ]$ then passed to the 3D ConvLSTM with 2 hidden layers for motion learning. The LSTM hidden states at the second level with a shape of [M/2, H/4, W/4, 128] are then fed through a shallow 3D CNN with two heads to produce the parameters of the prior distribution with shape [M/2, H/4, W/4, 16] that are later sampled to produce the latent tensor Z. This latent tensor is then concatenated with the hidden states and finally propagated through the decoder (3D-DEC) to predict the frames H timesteps into the future. The model is applied recursively by using the newly generated frame as input if the ground truth is not available. Specifically, if the frames are observed up to time C, then the model will use the ground truth as input up till time C-M then a combination of the ground truth and predicted frames between time C-M to C, and then finally, only the predicted frames as input from time C onwards. During training, a separate 3D encoder (not shown in Figure \ref{fig:architecture}) is used to generate the posterior distribution to optimize the KL divergence. 


\begin{figure}
	\begin{center}
		\begin{tabular}{c}			\includegraphics[width=1\columnwidth]{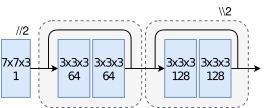}\\
			(a) 2-block 3D ResNet-18 encoder\\ \includegraphics[width=1\columnwidth]{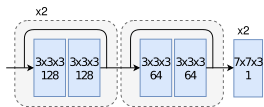}\\
			(b) 2-block 3D ResNet-18 decoder
		\end{tabular}
		\caption{Architecture of our encoder and decoder. The encoder outputs a 4D tensor with a spatial resolution of (H/4,W/4). Exception the decoder, each filter output is followed by a 3D batch-norm and ReLU. An downsampling operation with stride 2 is indicated by "//2" and an upsampling with stride 2 by "x2".}
		\label{fig:encoderdecoder}
	\end{center}
\end{figure}

As shown in Figure \ref{fig:encoderdecoder}, we use a 2-block 3D ResNet-18 for both the encoder and decoder in contrast to \cite{hierarchyvrnn} that use a full ResNet. We find this to be sufficient especially since the 3D ConvLSTM itself serve as an extension of the 3D CNN for learning complex spatiotemporal features given a window of \textbf{M} frames. 
Furthermore, by truncating the number of blocks to 2, we reduce the total number of parameters significantly which allow us to devote additional resources to our 3D ConvLSTM with 128 hidden units that contains 7m parameters per level. However, we also want to avoid the other end of not having a feature extractor at all \cite{wang-predrnn,wang-e3d} since they have been shown to extract useful features that tend to be task specific at the higher blocks and more general purpose at the lower blocks. In short, we propose to use a smaller feature extractor CNN and a larger LSTM. We show in our experiments that modelling the architecture in such a manner allows us to outperform the state-of-the-art while requiring fewer parameters.

\section{Experiments}
\label{sec:experiments}
\begin{figure*}[t]
\begin{center}
    \setlength{\tabcolsep}{2pt}
    \begin{tabular}{rcc}
    \raisebox{1\height}{\footnotesize Ground Truth}&\includegraphics[width=0.8\textwidth]{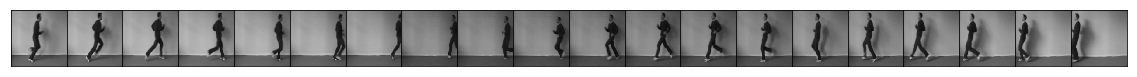}\\[-0.25cm]
    \raisebox{1\height}{\footnotesize E3D LSTM}&\includegraphics[width=0.8\textwidth]{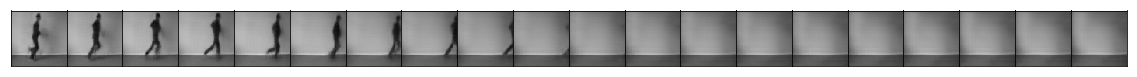}\\[-0.25cm]
    \raisebox{1\height}{\footnotesize Ours}&\includegraphics[width=0.8\textwidth]{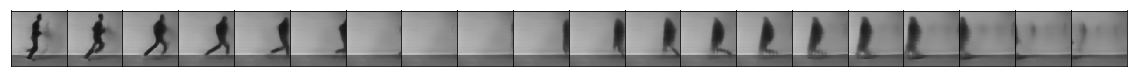}\\
    \raisebox{1\height}{\footnotesize Ground Truth}&\includegraphics[width=0.8\textwidth]{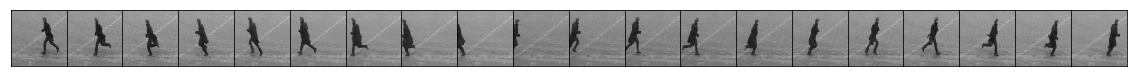}\\[-0.25cm]
    \raisebox{1\height}{\footnotesize E3D LSTM}&\includegraphics[width=0.8\textwidth]{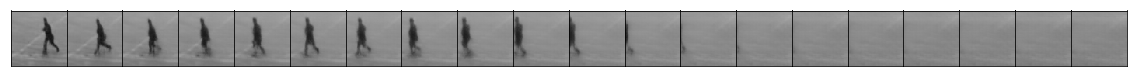}\\[-0.25cm]
    \raisebox{1\height}{\footnotesize Ours}&\includegraphics[width=0.8\textwidth]{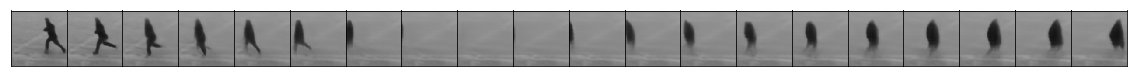}
    \end{tabular}
\caption{Prediction on the KTH action dataset. Our method recovers the disappearing man.}
\label{fig:qualitativekth}
\end{center}
\end{figure*}

\begin{figure*}[t]
\begin{center}
    \setlength{\tabcolsep}{2pt}
    \begin{tabular}{rcc}
    \raisebox{1\height}{\footnotesize Input}&\includegraphics[width=0.4\textwidth]{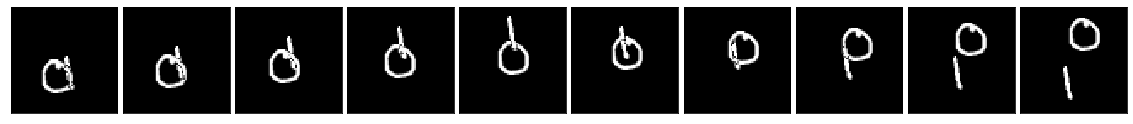}&\includegraphics[width=0.4\textwidth]{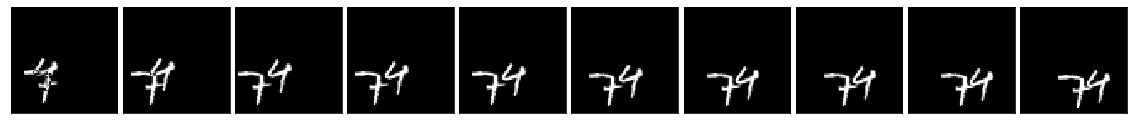}\\[-0.18cm]
    \raisebox{1\height}{\footnotesize GroundTruth}&\includegraphics[width=0.4\textwidth]{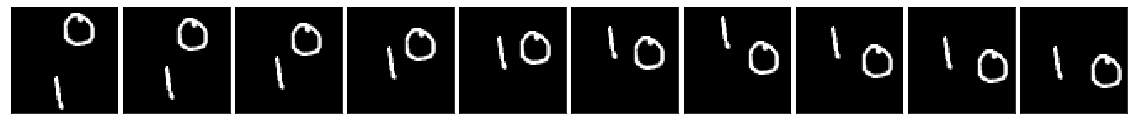}& \includegraphics[width=0.4\textwidth]{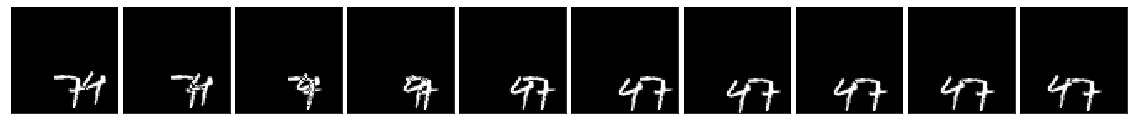}\\[-0.18cm]
    \raisebox{1\height}{\footnotesize ConvLSTM}&\includegraphics[width=0.4\textwidth]{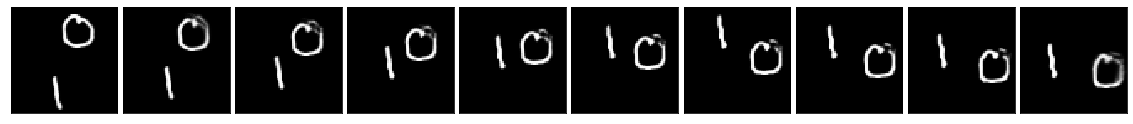}& \includegraphics[width=0.4\textwidth]{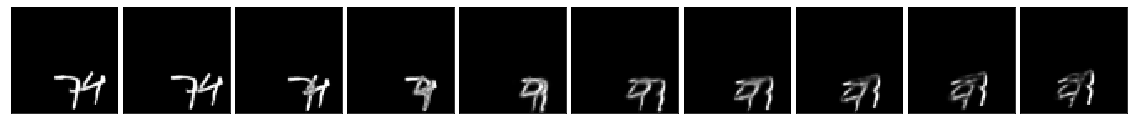}\\[-0.18cm]
    \raisebox{1\height}{\footnotesize VarConvLSTM}&\includegraphics[width=0.4\textwidth]{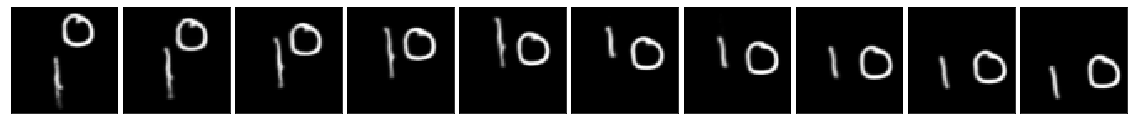}& \includegraphics[width=0.4\textwidth]{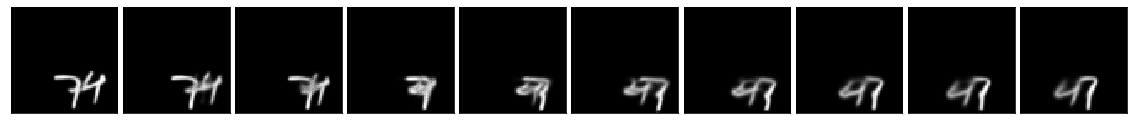}\\[-0.18cm]
    \raisebox{1\height}{\footnotesize E3DLSTM}&\includegraphics[width=0.4\textwidth]{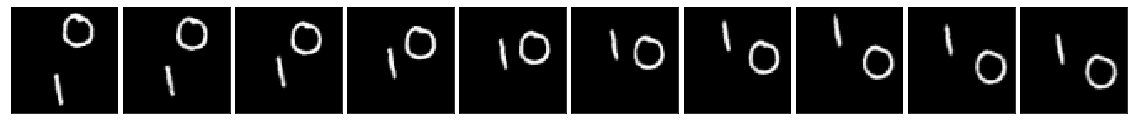}& \includegraphics[width=0.4\textwidth]{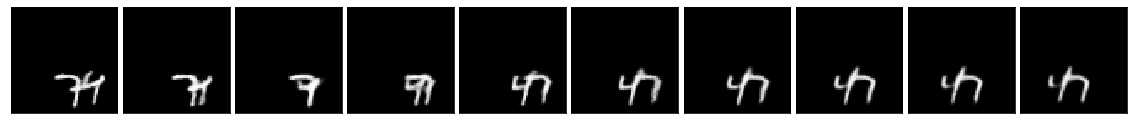}\\[-0.18cm]
    \raisebox{1\height}{\footnotesize Ours} &\includegraphics[width=0.4\textwidth]{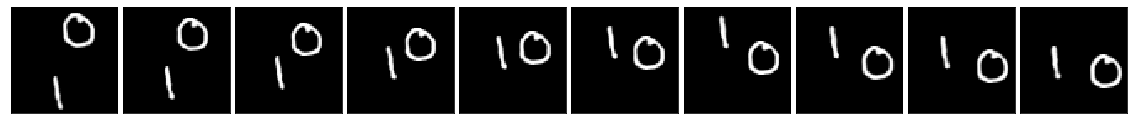}& \includegraphics[width=0.4\textwidth]{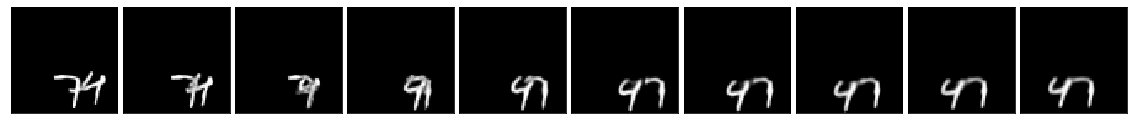}\\
    \end{tabular}
\caption{Prediction on the moving MNIST dataset. We obtain visually pleasing results even on complex example shown in right. Furthermore, our results indicate less blur.}
\label{fig:qualitativemnist}
\end{center}
\end{figure*}

\subsection{Comparison to state-of-the-art}
\label{sec.soa}
We compare our approach to several state-of-the-art methods using publicly available source code and model where available with default parameters and using standard metrics such as frame-wise Mean Squared Error (MSE), Structural Similarity Index (SSIM) and Peak Signal to Noise Ratio (PSNR). We sample 50 predictions from the stochastic models for each ground truth test sequence and average the metrics across the test set. Note that sampling is done only for the purpose of evaluation in order to get the average performance and is not required for deployment. \\

%
%

\noindent
\textbf{Training Details:}
We initialize the weights of our truncated 3D ResNet-18 encoder and decoder with weights pre-trained on the Kinetics-400 dataset \cite{kinetics} and all other components using PyTorch's default initializer.
We use the Adam optimizer with default hyperparameters, a learning rate of $10^{-3}$ with no weight decay, a batch size of 6 and the L1+L2 reconstruction loss that was also used in \cite{wang-predrnn,wang-e3d}. We train the model using beta warm-up \cite{betawarmup} and have it gradually predict into the future using its own predictions as input \cite{ss}. \\

\noindent
\textbf{The Moving MNIST dataset}
\cite{srivastava} consists of two digits (0 to 9) of size 28 x 28 moving inside a 64 x 64 patch. 
The digits are chosen randomly from the MNIST training set and placed at random locations inside the patch. 
Each digit is assigned a velocity whose direction is chosen uniformly at random on a unit circle and whose magnitude is also chosen uniformly at random over a fixed range. 
The digits bounce off the edges of the 64 x 64 frame and overlap as they move past each other. The training set contains 10,000 sequences while the validation and test sets 1,000 sequences each. By default, the sequences are all 20 frames long and the models are trained to predict the next 10 frames given the first 5 or 10 as input.

Table \ref{table:quantitativemnist} shows the performance of the models when using 5 frames to predict 10 and 15 frames into the future and when using 10 frames to predict 10 and 20 frames into the future. Our method demonstrates its promise, outperforming both the state-of-the-art deterministic (E3D-LSTM \cite{wang-e3d}) and stochastic (VRNN \cite{vrnn}) models, with the latter by a large margin. We also improve over the E3D-LSTM \cite{wang-e3d} despite having fewer parameters. 
We were only able to train the smallest variant of the models presented in \cite{hierarchyvrnn} which nevertheless contains 62 million parameters. Interestingly, we also outperform them. These results thus indicate the impact of our new design and the novel latent loss.
\begin{table*}[t]
\begin{center}
\resizebox{1\textwidth}{!}{
\begin{tabular}{|l|l|cc|cc|cc|cc|r|}
\hline
\multirow{2}{*}{Type} & \multirow{2}{*}{Model} &
\multicolumn{2}{c|}{$\text{x}_{1:5} \rightarrow \hat{\text{x}}_{6:15}$} & \multicolumn{2}{c|}{$\text{x}_{1:5} \rightarrow \hat{\text{x}}_{6:20}$} & \multicolumn{2}{c|}{$\text{x}_{1:10} \rightarrow \hat{\text{x}}_{11:20}$} & \multicolumn{2}{c|}{$\text{x}_{1:10} \rightarrow \hat{\text{x}}_{11:30}$} & \multirow{2}{*}{\# Params} \\
 &  & SSIM & MSE & SSIM & MSE & SSIM & MSE & SSIM & MSE & \\
\hline
\multirow{3}{*}{Deterministic} \ & 2D ConvLSTM \cite{convlstm} & 0.662 & 111.1 & 0.482 & 154.3 & 0.763 & 82.2 & 0.660 & 112.3 & \textbf{2.8M} \ \ \\ 
& PredRNN++ \cite{wang-predrnn} & 0.793 & 66.2 & 0.769 & 79.2 & 0.870 & 47.9 & 0.821 & 57.7 & 15.4M \ \ \\
& E3D-LSTM \cite{wang-e3d} & 0.853 & 53.4 & 0.801 & 64.1 & \textbf{0.910} & 41.3 & 0.872 & 47.6 & 38.7M \ \ \\
\hline
\multirow{2}{*}{Stochastic} & Variational 2D ConvLSTM \cite{vrnn} & 0.733 & 91.1 & 0.564 & 126.4 & 0.816 & 60.7 & 0.773 & 83.5 & 2.9M \ \ \\
& Improved VRNN \cite{hierarchyvrnn} & 0.772 & 123.1 & 0.728 & 162.2 & 0.776 & 129.2 & 0.699 & 194.3 & 62M \ \ \\
& Variational 3D ConvLSTM (Ours) & \textbf{0.864} & \textbf{51.4} & \textbf{0.805} & \textbf{63.2} & 0.896 & \textbf{39.4} & \textbf{0.874} & \textbf{47.54} & 12.9M \ \ \\
\hline
\end{tabular}
}
\end{center}
\caption{Results on the Moving MNIST dataset when using 5 frames to predict 10 ($\text{x}_{1:5} \rightarrow \hat{\text{x}}_{6:15}$) and 15 ($\text{x}_{1:5} \rightarrow \hat{\text{x}}_{6:20}$) frames into the future, and when using 10 frames to predict 10 ($\text{x}_{1:10} \rightarrow \hat{\text{x}}_{11:20}$) and 20 ($\text{x}_{1:10} \rightarrow \hat{\text{x}}_{11:30}$) frames into the future. The metrics are computed frame-wise. Higher SSIM or lower MSE scores indicate better results. Finally, the rightmost column indicate the number of parameters for the various models.}
\label{table:quantitativemnist}
\end{table*}

\begin{table*}
\centering
\setlength\tabcolsep{4pt}
    \begin{tabular}{@{} lr @{}}
\small
\begin{tabular}{@{}|l|cc|cc|cc|@{}}
    \hline
\multirow{2}{*}{Model} & \multicolumn{2}{c|}{$\text{x}_{1:10} \rightarrow \hat{\text{x}}_{11:30}$} & \multicolumn{2}{c|}{$\text{x}_{1:10} \rightarrow \hat{\text{x}}_{11:50}$} & \multicolumn{2}{c|}{$\text{x}_{1:10} \rightarrow \hat{\text{x}}_{11:70}$} \\
    & SSIM & PSNR & SSIM & PSNR & SSIM & PSNR \\
    \hline
2D ConvLSTM \cite{convlstm} & 0.712 & 23.58 & 0.639 & 22.85 & 0.551 & 20.13 \\
    PredRNN++ \cite{wang-predrnn} & 0.865 & 28.47 & 0.741 & 25.21 & 0.702 & 23.51 \\
E3D-LSTM \cite{wang-e3d} & \textbf{0.879} & \textbf{29.31} & 0.810 & 27.24 & 0.798 & 26.82 \\
    \hline
Variational 2D ConvLSTM \cite{vrnn} & 0.787 & 25.76 & 0.733 & 24.83 & 0.672 & 23.13 \\
Variational 3D ConvLSTM (Ours) & 0.866 & 28.31 & \textbf{0.852} & \textbf{27.89} & \textbf{0.846} & \textbf{27.66} \\
    \hline
\end{tabular}
    &
\includegraphics[width=0.34\linewidth,valign=c]{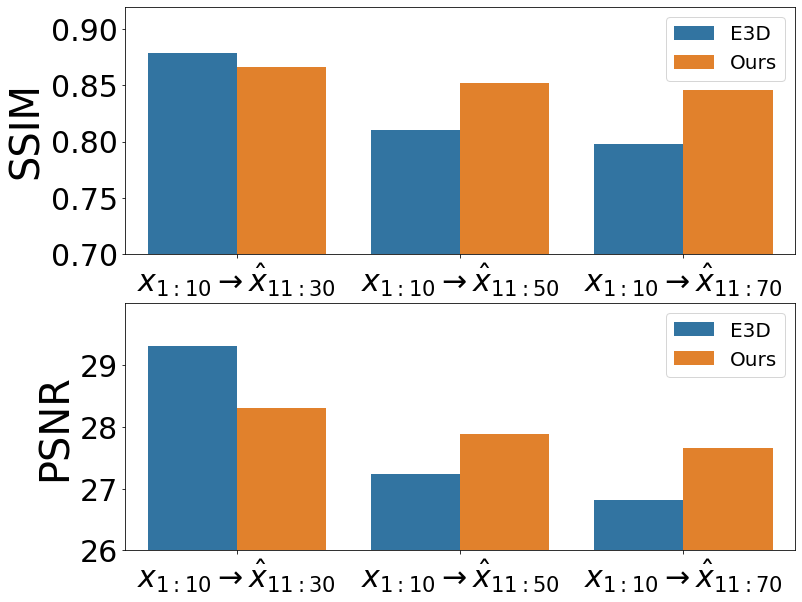}
\end{tabular}
\vspace{0.05cm}
\caption{Results on the KTH action dataset when using 10 frames to predict 20 ($\text{x}_{1:10} \rightarrow \hat{\text{x}}_{11:30}$), 40 ($\text{x}_{1:10} \rightarrow \hat{\text{x}}_{11:50}$), and 60 ($\text{x}_{1:10} \rightarrow \hat{\text{x}}_{11:70}$) time steps into the future. The metrics are computed frame-wise. Higher SSIM and PSNR scores indicate better results. The bar chart on the right highlights the difference between our model and the E3D-LSTM. Our model performs much better for longer predictions.}
\label{table:quantitativekth}
\end{table*}

We present some visual results in Figure \ref{fig:qualitativemnist} where the first row illustrates the input sequence $\textbf{x}_{1:10}$, the second the ground truth for the predicted sequence $\textbf{x}_{11:20}$ and all subsequent rows the predictions made by the various models $\hat{\textbf{x}}_{11:20}$. 
It can firstly be seen that the injection of stochasticity causes the Variational ConvLSTM to output predictions that are blurrier than its deterministic counterpart. 
This could principally be due to the fact that information coming from the latent nodes act as noise and thus interferes with reconstruction. 
Unlike the ConvLSTM however, the digits generated by the Variational ConvLSTM are closer to the ground truth resulting in a performance that is generally superior as evidenced by the quantitative scores in Table \ref{table:quantitativemnist}. We can then observe our model producing the best results. 
This signals that the blurry reconstructions manifested by the Variational ConvLSTM are counteracted by replacing all 2D convolutions with 3D. 
Intuitively, our novel loss also acts as a stronger regularizer for the reconstructions.\\

\noindent
\textbf{The KTH dataset}
\cite{kth} consists of humans performing 6 types of actions: boxing, clapping, waving, jogging, running, and walking under 4 scenarios: outdoors, outdoors with scale variation, outdoor with different clothes, and indoors with a homogeneous and static background. Each video is recorded at 25 fps and lasts an average of 4 seconds. We follow the experimental setup in \cite{villegas} using persons 1-16 for training and 17-25 for testing and resize each frame to 128x128 pixels. The models are trained to predict the next 10 frames given the first 10 as input.
Table \ref{table:quantitativekth} presents the performance of the various models when predicting the next 20, 40 and 60 frames. It can be observed that our model lags slightly behind the E3D-LSTM when predicting short term but performs much better when tasked to predict further into the future. This difference is highlighted on the bar chart beside Table \ref{table:quantitativekth} that shows the performance of our model degrading at a slower rate than the E3D-LSTM.
The results can be explained by Figure \ref{fig:qualitativekth} where each row represents the output sequence $\hat{\textbf{x}}_{11:50}$ spaced 2 frames apart. 
It can be observed from the figures that the predictions coming from our model are blurrier than E3D, resulting in metrics that are inferior although we make up for it by being able to predict the individual re-entering the scene.
The quantitative scores also show that our variational method is much better than the Variational 2D ConvLSTM \cite{vrnn}. 
These findings once again demonstrate the effectiveness of our architecture for modelling spatiotemporal data.

\begin{table*}
	\begin{center}
	    \resizebox{1\textwidth}{!}{
		\begin{tabular}{|l|cc|cc|cc|cc|r|}
			\hline
			\multirow{2}{*}{Model} & 
			\multicolumn{2}{c|}{$\text{x}_{1:5} \rightarrow \hat{\text{x}}_{6:15}$} & 
			\multicolumn{2}{c|}{$\text{x}_{1:5} \rightarrow \hat{\text{x}}_{6:20}$} & 
			\multicolumn{2}{c|}{$\text{x}_{1:10} \rightarrow \hat{\text{x}}_{11:20}$} & 
			\multicolumn{2}{c|}{$\text{x}_{1:10} \rightarrow \hat{\text{x}}_{11:30}$} & 
			\multirow{2}{*}{\# Params}\\
			& SSIM & MSE 
			& SSIM & MSE 
			& SSIM & MSE 
			& SSIM & MSE &\\
			\hline
			2D ConvLSTM \cite{convlstm} & 0.662 & 111.1 & 0.482 & 154.3 & 0.763 & 82.2 & 0.660 & 112.3 & \textbf{2.8M} \ \ \ \\
			Variational 2D ConvLSTM \cite{vrnn} & 0.733 & 91.1 & 0.564 & 126.4 & 0.816 & 60.7 & 0.773 & 83.5 & 2.9M \ \ \ \\
			Variational 3D ConvLSTM & 0.857 & 52.1 & 0.797 & 63.8 & 0.887 & 41.8 & 0.868 & 49.7 & 12.9M \ \ \ \\
			Variational 3D ConvLSTM + LL Criterion (ours) & \textbf{0.864} & \textbf{51.4} & \textbf{0.805} & \textbf{63.2} & \textbf{0.896} & \textbf{39.4} & \textbf{0.874} & \textbf{47.5} & 12.9M \ \ \ \\
			\hline
		\end{tabular}
	    }
	\end{center}
	\caption{Ablation study on the Moving MNIST dataset. The metrics are computed frame-wise. Higher SSIM or lower MSE scores indicate better results.}
	\label{tab:ablationmnist-components}
\end{table*}

\begin{figure*}[t]
	\scriptsize
	\begin{center}
		\begin{tabular}{cc}
			\includegraphics[width=0.41\textwidth]{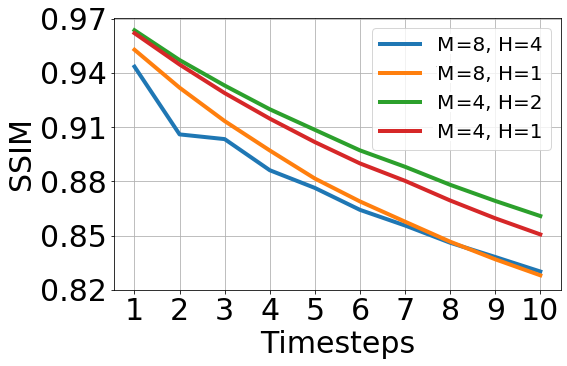}& \includegraphics[width=0.38\textwidth]{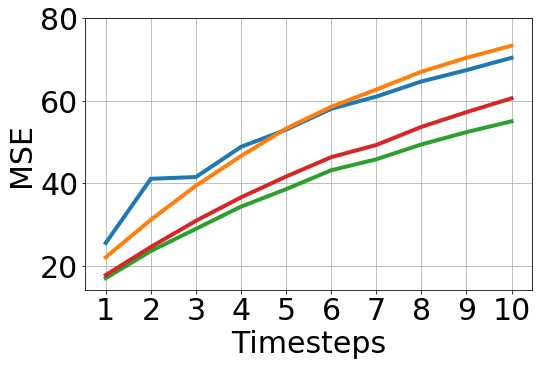}\\
			\includegraphics[width=0.41\textwidth]{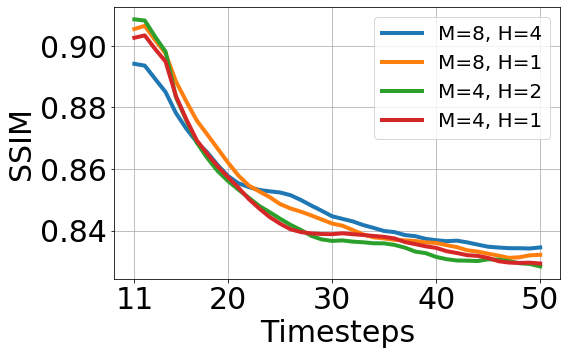}& \includegraphics[width=0.38\textwidth]{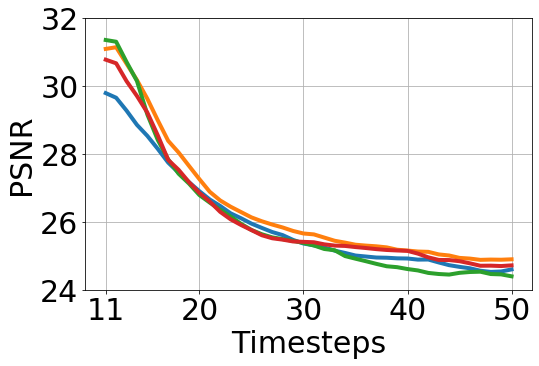}\\
		\end{tabular}
		\caption{The effect of \textbf{M} window size and the output horizon $\mathcal{H}$ on the performance. The first row shows the SSIM and MSE scores on the moving MNIST action dataset, while the second row the SSIM and PSNR on the KTH action dataset.}
		\label{fig:ablationkth-inputoutput}
	\end{center}
\end{figure*}

\subsection{Ablation Study}\label{sec:ablationstudies}
\textbf{Effectiveness of each component:}
We quantize the effect of each contribution in Table \ref{tab:ablationmnist-components} on the moving MNIST dataset.
It can be seen that the introduction of stochasticity into the recurrent network allows it to better tackle uncertainties in the recurrent dynamics which results in better predictions, significantly lowering the MSE from 82.2 to 60.7. 
Additionally, swapping out the 2D convolutions in place for 3D brings about  significant improvements to the model, lowering the MSE by approximately another 20 points. 
This makes sense since 3D convolutions operate on both the spatial and temporal axis, letting the architecture capture relationships in said dimensions. 
Finally, it is quite apparent that the introduction of the log-likelihood criterion has a noticeable effect, further bringing down the MSE by approximately 2 points.
This can be interpreted in various ways: (1) that the resulting loss function empirically helps the network traverse towards a better local minima and (2) that the added regularizer helps the recurrent model strengthen its ability at expressing complex distributions. 
Intuitively, appending the log-likelihood criterion to the KL divergence has some conceptual similarity to the use of the L1+L2 loss functions that has been empirically shown in \cite{wang-predrnn,wang-e3d} to be better than the individual counterparts. 
In conclusion, each component brings a definitive upgrade to the model and together, lend it the advantage it needs to outperform the deterministic and stochastic state-of-the-art models~\cite{vrnn,wang-e3d,wang-predrnn,hierarchyvrnn}.

\noindent
\textbf{Window size and output horizon:}
Recall from section \ref{sec:convolutionalrecurrentnetworks} that the advantages 3D convolutions have over its 2D counterpart when paired with an LSTM is that (1) the LSTM state transitions are no longer completely reliant on its hidden states for motion information and (2) that one could vary the window size (\textbf{M}) and the output horizon ($\mathcal{H}$) at each timestep without having to change the architecture. In this experiment, we conduct additional studies on our model where we varied \textbf{M} and $\mathcal{H}$, the number of input frames and the output horizon respectively at each timestep to study the effects different design choices have on our model. 
As expected, the plots in Figure \ref{fig:ablationkth-inputoutput} show a degradation in the metrics over time regardless of the configuration in use. The KTH plots exhibits an exponential decay whereas the moving MNIST is more linear. The plots also show that a smaller window size is better for the MNIST dataset but has no clear difference for the KTH action dataset. 
Interestingly, there are cases that favour a large window size and output horizon and some other cases, that do not.
For longer predictions however, the plots show that the various configurations are quite similar in terms of performance.

\section{Conclusion}
\label{sec:conclusion}
We have presented a deep neural network for future frame prediction that performs well at predicting long term. 
Our method uses 3D convolutions throughout the entire architecture and is trained using a latent loss that includes a \emph{specific log-likelihood criterion}. 
Theoretically and experimentally, we have shown the effects of these two contributions. First, the use of latent random variables in a 3D recurrent model enables it to persistently generate predictions well beyond the time steps it was trained for and second, the log-likelihood criterion helps direct the model towards a better solution without an increase in model complexity. 
This model outperforms prior stochastic methods by a good margin while obtaining a performance that is on-par with the state-of-the-art deterministic models for short-term future frame prediction while being superior when generating even further into the future. 

We also proposed the benefits of using a smaller convolutional network for encoding and decoding videos as it alleviates the burden on the ConvLSTM to jointly learn short-term spatiotemporal features and long-term dynamics, while continuing to keep the total number of parameters manageable. We use a truncated 3D ResNet-18 (2 blocks) which reduces the total number of parameters by 60 million as the low-level spatiotemporal features captured by the first few blocks of the 3D ResNet-18 are sufficient for the ConvLSTM to further learn. We believe these findings are useful for the development of more efficient and effective spatiotemporal models with variational recurrent architectures.

\noindent
\textbf{Acknowledgment}
\noindent
This research/project is supported by the National Research Foundation, Singapore under its AI Singapore Programme (AISG Award No: AISG-RP-2019-010).
{\small
\bibliographystyle{ieee_fullname}
\bibliography{egbib}
}

\end{document}